# From Representation to Enactment: The ABC Framework of the Translating Mind

**Michael Carl *, Takanori Mizowaki [1], Aishvarya Raj [2], Masaru Yamada [1], Devi Sri Bandaru *, Yuxiang Wei * and Xinyue REN [3]**

* **Center for Research and Innovation in Translation and Translation Technology, Kent State University, USA and**
[1] Rikkyo University, Tokyo,
[2] Independent researcher, London,
[3] City University of Hong Kong, Hong Kong,

**Abstract**

Building on the third-wave Extended Mind (EM) theory and radical enactivism, this article suggests an alternative to representation-based models of the mind. We build on the ABC framework, in which translation is not understood as the manipulation of static interlingual correspondences but an enacted activity, dynamically integrating affective-evaluative, behavioral-enacting, and cognitive-inferential (ABC) processes. Drawing on affordance theory, we argue that the translating mind is the dynamically organized process through which the translator and its environment constitute and transform a landscape of enlanguaged translation affordances for action, that emerges through loops of brain–body–environment interactions. This non-representational account reframes translation as skillful participation in sociocultural practice, where meaning is co-created in real time through embodied interaction with texts, tools, and contexts.

## 1. Introduction

Classical internalism argues that cognition involves manipulating internal representations that stand for external state of affairs, where mental processes trigger actions that modify the world. In the functionalist view, however, the mind is defined by functional patterns rather than the location of the processes. Under this view, external tools can perform cognitive functions, which can become genuine parts of cognitive systems. Thus, some extended mind (EM) theorists (e.g., [1]) argue that if external tools fill the equivalent functional roles as internal processes, they are on a par with the internal elements.

EM hypotheses have been identified in three evolving 'waves [2]. The initial wave is parity-driven, primarily focusing on the functional similarities between external and internal elements [1]. In other words, external resources bearing the equivalent functional roles should be treated as part of the mind. The second wave of EM [3,4] shifted the focus to complementarity principles, highlighting cooperative and transformative roles of external elements. External resources act as scaffolds that shape, extend, and sometimes constitute cognitive processes, depending on the degree of integration with the organism. Clark's later works (e.g., [5,6]) edge towards this second wave of EM emphasizing active inference model [6], in which cognitive processes are primarily brain-based, while the mind may incorporate external elements as scaffolds or tools. The first two EM waves view external and internal elements in terms of their functional characteristics. As we will lay out in section 3, some enactivists reject these EM frameworks and push the discussion into a third version.

Kirchhoff & Kiverstein [2] propose a radical enactive perspective that conceptualizes the mind as distributed across brain, body, and environment, in which the environment can play a constitutive role rather than a mere causal aid (scaffolds). Here, the brain, body, and environment form a single, integrated system where the boundaries of the mind are fluid and context dependent. In this view, the environment is not a passive backdrop but it can constitutively contribute to the generation of meaning. It posits a mind-cognition continuum, seeing the mind as a system collectively constituted by embodied, embedded, enactive, extended, and affective (4EA) processes.

Building on radical enactivism, in section 2 we develop a novel ABC framework of mind [7], addressing affective-evaluative (feeling/emotion-related), behavioral-enactment (action/observation-related), and cognitive-inferential (reflection/thought-related) processes. Behavioral enactment patterns are embodied skills that established coupling patterns with the environment. Cognitive-inferential activity is the process of active sense-making that is called upon when automated behavioral routines turn out to be insufficient, while the affective-evaluative dimension provides an evaluative aspect of all sense-making.

The term "cognition" is often used in a broad, umbrella sense that includes all types of information processing and input-output transformations, such as memory, attention, reasoning, perception, etc. - essentially all mental processes. However, the term "cognition" can also be opposed to other mental processes, e.g., affective and behavioral, as in the ABC framework. Cognition in this narrower sense is then related to deliberate or effortful mental processing, problem-solving, reasoning, or explicit learning. We refer to this narrow sense as cognitive-inferential or inferential cognition, as opposed to, e.g., sensorimotor cognition.

The ABC framework is related to dual-process models. Some dual-process models differentiate between cognitive (i.e. cognitive-inferential) and intuitive/affective processes. Kahneman [8] distinguishes intuitive (System 1) from analytical (System 2) processes. Evans & Stanovich [9] propose an interaction model of affective and cognitive systems in decision-making, while Ajzen's [10] Theory of Planned Behavior links attitudes to behavioral intentions and actions, integrating subjective norms and perceived control. We integrate components of these process models suggesting a three-tier ABC model.

Here, the mind is the emergent organization of ABC dynamics; it emerges from the self-organizing coordination of affective-evaluative, behavioral-enacting, and cognitive-inferential dynamics within an organism's ongoing coupling with its environment. This view alleviates the need to decide what "extends" because the mind emerges dynamically with the processes that it includes.

In Cognitive Translation Studies (CTS), recent theoretical frameworks have also moved towards a more holistic view, recognizing that emotion and context are integral to mental processes of translation [11,12]. The translator's dynamic interaction with texts and the environment defines the extended mind, which includes the body, and surroundings. These situated cognition models of translation incorporate social context [e.g., 13] to account for both mental and environmental aspects of translation processing within socio-cognitive frameworks, where human cognition is considered fundamentally social and relational, and where the boundaries between the individual mind and the social environment become blurred. Nevertheless, representational assumptions remain pervasive in the relevant literature, where environmental elements are typically conceived as being represented by internal symbols or mental models that guide target-language production.

Muñoz and Rojo López [14], for instance, two of the key 4EA protagonists in CTS state that "Language units first activate representations in our minds, corresponding to their visual or auditory characteristics ...”. This statement suggests a weak embodiment assuming a modality-specific representational framework, similar to Barsalou’s [15] perceptual symbol systems which is also the basic tenor in [16]. While classical cognitivism prioritizes amodal representations, the notion of modal representation moves away from old ways of representational thinking but often get trapped by the very language and models they are trying to replace. This creates an unease which is exemplified by Sannholm and Risku [13] when they say that “Contrary to its predecessors in cognitive science, situated cognition does not describe representations as units that actually *re*present ‘reality’ ... but rather as patterns generated by one’s own actions (without any *re*presentational quality or necessity)” (italics in the original). However, if there are no *re*presentational qualities whatsoever, retaining the term seems misleading.

In this article, we suggest a non-representational view which rejects the assumption that representations (modal or amodal) function as basic cognitive units. Instead, a translator, we claim, enacts translations through situated, embodied, and affectively modulated interaction with text, tools, and context. On this view, meaning emerges from the partially constitutive coupling of ABC dynamics with the environment and enlanguaged affordances. This view circumvents the apparent paradox that "meaning happens in our heads, and only in our heads. It never leaves" while simultaneously “meaning is a social phenomenon” as [14] suggest.

To overcome this theoretical gap and escape the vocabulary trap of classical cognitivism, we move beyond representationalism entirely. We deploy the concept of "enlanguaged affordances" [17, 18] for translation, which was specifically developed to replace the need for mental representations when explaining how humans perform high-level linguistic and abstract thinking. Affordances are possibilities for action provided by the environment, i.e., relations between aspects of the material environment and abilities available in a form of life (such as a translator). Enlanguaged affordances are the direct action possibilities and invitations for abstract or linguistic thinking within a language-saturated environment.

Theorists of affordances argue that the brain does not need to construct an internal model of the world. Instead, the explanatory work traditionally assigned to representations is shifted entirely to the environment and the relation to it. Because the human ecological niche is saturated with sociomaterial practices (such as speech, text, symbols, and cultural norms), humans can navigate complex thoughts by directly interacting with these environmental invitations. Rather than mapping external stimuli to internal representations, translation in this view, is a process of skillful organization and participation in a sociocultural practice, where text comprehension and production are jointly realized in the translator’s embodied engagement with their environment. This view is supported by neuroscience which can provide evidence that supports the understanding that the human mind's primary function is to optimize new behaviors, rather than represent the environment.

This ecological perspective can be mathematically and structurally modelled within an active inference framework. Active inference formally defines the boundary of an autonomous system (such as a cell, organism, or, as in our case, a translator) through Markov blanket (MB), separating internal states from external states via sensory and active states. While the components of the ABC architecture and its computational modelling are thus well-established in CTS and neighboring disciplines, this paper provides a conceptual synthesis and integrates those ingredients into one dynamical architecture for translation, treating affect, behaviour and cognition as different temporal/functional layers of one coupled dynamical system.

## 2. Three Waves of the Extended Mind

Kirchhoff & Kiverstein [2] distinguish between successive "waves" of thinking about the extended mind and develop an enactive third EM wave that draws on enactivism. Our ABC framework will be based on their third wave EM perspective.

*2.1 Extended Mind (EM) Theories and the Enactivist Perspective*

According to [2], first wave EM theorists recognize that external artifacts can be constitutive of cognition when functionally integrated. In their view, the mind is in the head, which can be extended with external resources. Clark & Chalmers [1] use the coupling argument which states that when a system is tightly and reliably coupled with an external element (e.g., a translator with a CAT tool, or Otto with a notebook and a GPS), that element becomes part of the cognitive system itself. The internal and external elements function as a unitary cognitive process.

Second wave EM suggests that external resources such as tools, language, and cultural practices are complementary to the organic parts of the mind, as they actively participate in shaping and constituting how these enactments unfold. In line with the second wave, Littau [20] posits that "Media are not merely instruments with which writers or translators produce meanings; rather, they *set the framework within which something like meaning becomes possible at all."* (emphasis in the original), seeing translation as materially scaffolded and socially embedded.

The third-wave EM framework conceptualizes the mind as unfolding through continuous loops of sensorimotor engagement with technological and cultural practices. The notion of sensorimotor contingencies [21] emphasizes that perception is not a passive registration of external "states of affairs," but an active process of inference about the causes of sensory signals. Since the actual causes of sensory input are conditioned on anticipated and past actions, perceptual inference concerns observations that are themselves generated through action of the observing agent. In this sense, the self-evidencing system can be described as "the author of its own sensations" [22].

For translation, this implies that translators navigate sensorimotor contingencies shaped by reading, writing, tool use, and socio-cultural expectations. Each "sensory signal" (a string of words, a fuzzy match, a client instruction) is interpreted in light of anticipated responses. Just as sensory inputs depend on action (what I do changes what I perceive), the translator's interpretive stance depends on their translational actions: consulting a dictionary, choosing a draft phrasing, testing a formulation in the target language (TT), or interacting with CAT tools. Each action reshapes the available "input."

Ramstead et al.'s [23] idea that the system is the "author of its own sensations" parallels how translators generate the very conditions under which they construct meaning. By segmenting the text, trying out equivalents, or by foregrounding certain stylistic features, they partially create the interpretive signals they then act upon.

*2.2 Dynamic Entanglement Thesis (DEUTS)*

The thesis of *Dynamic Entanglement* (DEUTS) is central to the third wave. DEUTS [2, 24, 25] describes how the mind becomes reciprocally coupled with features of the environment. The DEUTS argument comprises two core theses:

1. The Dynamic Entanglement Thesis: Sensorimotor contingencies unfold in a way that yields the dynamic entanglement of the brain, body, and world. This entanglement involves non-linear causal interactions among neural, bodily, and environmental elements, where sensory and motor channels enter into reciprocal causal influence based on the agent's interaction with the environment.
2. The Unique Temporal Signature Thesis: Sensorimotor contingencies are characterized by unique temporal signatures which entail that the brain cannot be unplugged from the body and the world while keeping the phenomenal character of experience fixed.

The mind emerges in this view as a pattern of patterns within the ongoing entanglement of brain, body, and world. It is not a substance but a relational property of the embodied system in action. This pattern is affectively and normatively structured, orienting the organism toward what matters and sustaining meaningful interaction with its environment. This view mirrors Ryle's [26] Concept of Mind as a set of capacities and patterns, logically different from the material–biological–social domain, but an emergent, relational property that exists only when those components are appropriately and dynamically coupled.

### *2.3 Enlanguaged Affordances*

DEUTS suggests that the boundaries of the mind are fragile and continuously negotiated through participation in shared socio-cultural frameworks. Kirchhoff and Kiverstein argue that when a human is deeply coupled with an environmental tool (e.g., a translator interacting with a CAT tool, dictionary, or source text), the interaction operates on a unique temporal signature. As one cannot isolate where the "brain work" stops and the "tool work" begins, the boundaries between the brain, body, and world completely dissolve. The system exhibits feedback loops that are too fast and tightly integrated to be broken down into discrete "internal brain processing" vs. "external tool input".

This continuous, non-linear interaction provides the foundation for a third-wave account of the extended mind, which views the boundaries of the mind as fragile, temporary, and fundamentally open to negotiation within broader socio-cultural practices. The translator's mind extends not just because they look at a screen, but because they are synchronized with a wider network of linguistic rules and professional practices. By shifting the focus from static internal structures to dynamic, temporally entangled processes, the third wave allows us to bridge the gap between individual brain mechanics and the wider social environment. In the context of linguistic thought, this entanglement is operationalized through 'enlanguaged affordances,' [18, 28], action possibilities for spoken or written language.

Enlanguaged affordances exist as relational possibilities that become available through the coupling between an agent (e.g., a translator) and a historically structured linguistic and material environment. Enlanguaged affordances are neither internal representations nor fixed external properties. They are relational possibilities for perception, understanding, and action that arise through coupling with a historically and socioculturally structured environment, which are enacted at the interface between the currently constituted translating system and its environment. On this background, a translator can be conceptualized as an organism embedded in a language-saturated niche, directly attuning to the invitations of the text through a nested hierarchy of action-perception dispositions.

### *2.4 Relevance Realization and the optimal grip*

The human ecological niches are densely populated by socio-material practices (language, texts, and societal norms, etc.) in which individuals must develop robust mechanisms to navigate this complex landscape of enlanguaged affordances [2]. The source text, CAT tools, dictionaries, and socio-cultural norms constitute a massive, dense landscape of possibilities. This landscape is characterized by near-infinite semantic and syntactic possibilities, a purely algorithmic or representational search for meaning is intractable.

On a classical view, representations must explicitly track what is relevant and ignore what is not. In a complex world, updating a representational "world model" in real-time requires impossible computational power, known as the Frame Problem [29], which is also not solved (or even addressed [16]) in a modal representational system. Representations force a hard division between the subject (the mind holding the representation) and the object (the world being represented). This split makes grounding difficult when asking how symbols get their meaning [30]. If the answer is that they are grounded in modal or action-based representations, we might ask: How do those representations get their meaning, which leads to an infinite regress.

Relevance realization (RR, [27]) resolves this combinatorial problem by acting as a dynamic, self-organizing process through which the translator's cognitive system continuously reshapes its 'salience landscape'. Rather than retrieving a static internal map, the translator dynamically shifts their precision-weighting across the available socio-material field, allowing the most contextually appropriate options to stand out as immediate invitations for action. By classifying action-grounded dynamics as participatory and perspectival knowing[1], a translator is structurally coupled to a field of enlanguaged affordances. This interaction is transjective, direct, and unmediated, in which the emerging translation dynamically co-emerges from the reciprocal relationship between them.

RR is the real-time, non-representational optimization of the system's internal variational states, dynamically tuning the translator to harvest the most viable enlanguaged affordances offered in the current context. It is the process by which the translator's cognitive system continuously self-organizes to achieve an optimal grip [28] on the textual arena [27]. When reading a source text, certain linguistic structures, terms, or cultural nuances flash with high cognitive salience. This salience is not an internal representation; it is the immediate, felt awareness of a specific enlanguaged affordance.

### *2.5 Emergence of Propositions*

Relevance realization operates as a continuous, real-time optimization of a system's internal variational states. This process dynamically recalibrates the interpretive mechanism, allowing the system to actively seize the most viable enlanguaged affordances embedded within the immediate context. But how do truth-evaluable propositions emerge from this non-representational background?

Within a third-wave EM framework, propositions are not conceptualized as internal symbolic files retrieved from a mental database. Instead, they can be understood as the downstream product of 'socio-material scaffolding' [21]. The process begins with raw sensorimotor attunement to the text, which RR refines into a highly focused salience landscape. To stabilize this fleeting, experiential grip, the translator actively manipulates external linguistic artifacts—typing, deleting, and interacting with digital interfaces. This is not an internal mental state projected onto the screen; rather, it is a temporarily frozen invariant, an emerging proposition, which is co-constructed by the brain, body, and environmental tools. Propositions are thus externalized, socio-culturally constrained affordances that allow human organisms to manipulate abstract thought through direct material interaction, rendering traditional representational explanations entirely redundant.

A proposition is a temporarily stabilized, public linguistic structure. From our radical enactivism view, propositions emerge from enlanguaged affordances through a process of sedimented social practices, normativity, and skilled stabilizing practice. Instead of treating a proposition as an abstract, logical truth-value floating in a platonic realm or encoded in a brain database, this view argues that propositions are patterns of coordinated action stabilized by human language-games.

## 3. The ABC Layers of the Translating Mind

We build on the ABC framework of the translating mind [7] that categorizes translation processes into three functionally distinct, yet interdependent layers of affective-

[1] https://www.by-love-alone.com/blog/4-kinds-of-knowing

evaluative, behavioral-enacting, and cognitive-inferential dimensions (see Figure 1). The model provides a granular view of how the ongoing activity of sense-making unfolds on different timelines:

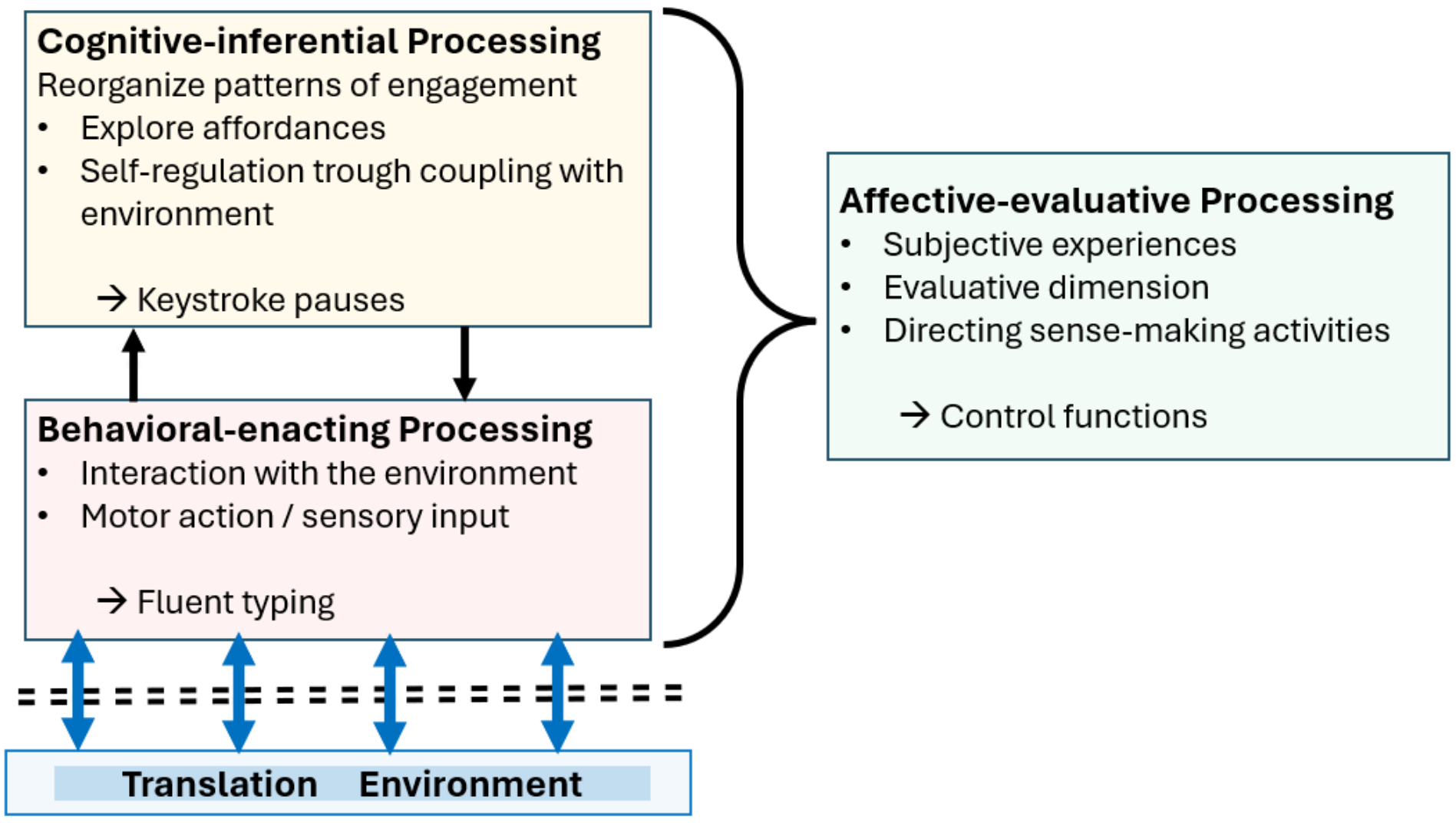

**Figure 1.** Relation in the ABC architecture and its relation to external environment.

- **Affective-evaluative processes** are feeling and emotion-related, providing an evaluative dimension that is crucial for successful agent-environment interaction. Affect and emotion are ways in which the organism enacts its concern-based engagement with the world. Grounded in embodied, biological attunement, they modulate how the organism couples with the environment, foregrounding what sustains its activity and shaping further sense-making.
- **Behavioral-enacting processes** encompass action- and sensation-related activities, including embodied skills and established behavioral routines. They manifest as observable actions—such as eye movements, typing, or gaze shifts—through which the translator engages with the environment. Through repeated practice, these sensorimotor couplings become consolidated into routinized cognitive-affective patterns that support fluent interaction and reduce the need for deliberative control.
- **Cognitive-inferential processes** are reflection and thought-related, involving active sense-making that becomes prominent when automatized behavioral patterns are insufficient to address the situation. This includes updating of internal states, memory, reasoning, and counterfactual simulation. Inferential cognition is functionally defined by its role in creating meaning, which can occur even at a pre-reflective level, especially when an organism needs to navigate uncertainty or resolve indeterminacy.

The different ABC processes can be conceptualized as nested hierarchy of variational internal states (see Figure 1). Internal states can be understood as embodied expectations that guide skilled action, the translator's structural coupling with the translation environment through action-perception loops. While representational accounts view behavioral patterns as encoded symbol routines, enactivist approaches see them as enacted dispositions of sensorimotor coupling. In translation, automated routines surface in process data as stretches of fluent typing or stable gaze trajectories.

### *3.1. The Behavioral-enacting Layer*

Because the behavioral-enacting layer is the point at which the translating system meets its environment. On its sensory side, this layer directly receives sensory input from the external states of the translation environment: gaze fixations take up visual information from the source and target texts, while on-screen changes, tool outputs, and other environmental cues are registered as they occur. On its active side, the same layer outputs the system's behaviour, enacting the motor actions (keystrokes, cursor movements, deletions, and gaze shifts) that alter the external states and thereby reshape the sensory input available at the next moment. Sensory and active states are therefore not a category separate from the behavioral layer; they are the two directions, reception and enactment, of the behavioral layer's coupling with the environment.

Gaze and keystrokes function as active states as they act on the environment, and they give rise to the sensory states that the system subsequently receives. When process data such as gaze fixations and keystrokes are described both as behavioral-layer phenomena and as sensory or active states, the two descriptions refer to the same events under complementary vocabularies: "behavioral layer" names the functional level within the ABC architecture, whereas "sensory" and "active" states name the formal role that those same behaviors. Within the nested ABC structure, the behavioral layer additionally possesses its own internal dynamics [36].

### 3.2. *The Cognitive-inferential Layer*

Cognitive-inferential states are here understood as deliberate reorganizations of ongoing sense-making that arise when habitual or automated translation routines no longer suffice. Rather than representing external truths, these states function as dispositions that stabilize and guide the unfolding translation process. They articulate expectations about which continuation will maintain coherence under the present conditions. They regulate self-organization through re-organization of the environmental coupling determining whether and how to:

- Maintain the current engagement patterns vs. reorganize
- Expand the environmental coupling (seek new resources) vs. consolidate existing patterns
- Shift between different modes of environmental interaction (analytical, intuitive, collaborative)

Cognitive-inferential states may be activated proactively—as anticipatory structuring of the translation flow—or reactively, when the translator encounters a disruption, such as when a chosen rendering no longer fits the emerging context.

### 3.3. *Anticipatory Cognitive-inferential States*

Anticipatory cognitive-inferential engagement operates at multiple reflective levels—ranging from predicting how a segment aligns with discourse or genre conventions, to foreseeing how readers, clients, or institutions may evaluate a choice, and recognizing when intended solutions conflict with contextual or stylistic norms. Such anticipation relies on metacognitive awareness of the reliability of one's expectations. Depending on expertise, task demands, and text difficulty, anticipations may involve:

- affordances of linguistic structures
- coherence emerging from translator–text interaction
- pragmatic effects of translation choices
- lexical and syntactic mappings across languages
- temporal dynamics of the translation process

### 3.4. *Affective-evaluative HORF states*

Whereas cognitive-inferential states can be described as anticipatory readiness-to-think and behavioral-enacting states as embodied readiness-to-act, affective-evaluative states are felt orientations that signal "this is workable," "this feels wrong," or "this matters." They are states of confidence, surprise, trust, curiosity, and similar affective tones that structure how cognitive-inferential and behavioral-enacting states unfold over time.

We currently distinguish between four affective states, Hesitation, Orientation, Revision, and Flow (HORF). HORF states do not represent truth but instead scaffold the direction of thought and action. They orient attention and behavior by providing signals that bias what seems relevant, urgent, or sufficient. Translators must be able to notice their own affective stance, modulate it, and sometimes realign it with task demands or audience expectations.

Affective-evaluative are closely tied to somatic awareness. According to Damasio's Somatic Marker Hypothesis [19], emotional processes are guided by bodily markers, which are physiological signals that shape decision-making. When facing a choice, somatic markers generate felt tendencies that steer a person toward advantageous options and away from disadvantageous ones. It requires proprioceptive sensitivity to one's own affective-motor readiness, i.e., the ability to register bodily responses to environmental situations, as well as interoceptive skills in detecting the emotional significance of situations.

#### 3.4.1. The Hesitation State

During a hesitation state, the translator experiences a moment of uncertainty in which predictions are assigned low precision. That is, the system registers doubt based on the current sensory or contextual evidence. Such hesitation often arises when lower layers (behavioral routines / sensory inputs) produce inconsistency due to noisy signals (e.g., conflicting input from terminological resources), or when discrepancies between expected and observed patterns become too large to be resolved automatically.

In a state of hesitation, automated processing temporarily suspends, keystrokes slow, gaze fixations linger, and the translator may pause reading, thus to avoid committing to a premature or error-prone choice. Simultaneously, the affective layer signals increased vigilance or mild discomfort, which may, in turn, encourage additional information-seeking or reinterpretation of the ST or TT. Decision-making is thus delayed until either the incoming data become more coherent or alternative cues raise the reliability of the available evidence.

Once prediction errors are reassessed, the system can transition out of hesitation. This transition may then take the form of a state of orientation, or a flow state. A hesitation thus serves as a self-regulating checkpoint: it is neither failure nor inertia but a protective recalibration mechanism ensuring that subsequent translation steps proceed under conditions of higher certainty and precision.

#### 3.4.2 The Orientation State

During the orientation state, the translator actively familiarizes herself with (a part of) the ST, scanning its structure, style, and potential difficulties. A state of orientation is more than simply reading; it calibrates the translator's perceptual and cognitive systems to the task at hand. As comprehension deepens, the precision of subsequent predictions for the most salient cues, terminology, and syntactic patterns, is increased, amplifying their impact on downstream processes. In this sense, orientation is not a passive stage, but a dynamic redistribution of attentional and affective resources: attention is selectively heightened for features judged relevant, while irrelevant noise is dampened.

A state of orientation thus constitutes a coordinated Affective–Behavioral–Cognitive reconfiguration: Affective states (interest, curiosity, tension, etc.) tune the sensitivity of perception and memory, behavioral routinized gazing patterns enact the process of exploration, and cognitive processes establish a provisional landscape of expectations about how the translation will unfold.

Taken together, these processes enable the translator to set the stage for later action in a Flow state, reducing uncertainty and increasing the efficiency of subsequent decision-making.

#### 3.4.3. The Revision State

During a revision state, the translator actively re-engages with already produced text, treating it as fresh sensory input. The translator reallocates attention and precision toward previously written target-text segments, signaling a readiness to detect discrepancies, improve fluency, or adjust meaning. This typically arises when the translator senses (often affectively) that something "doesn't fit" or when a better formulation becomes available.

At the behavioral layer, revision manifests gaze fixations on the TT, probably back through text, or increased gaze dwell time on earlier segments, deletions, retyping, or modifications. Automated routines give way to slower, more deliberate actions, enabling the translator to test alternative expressions and re-align the translation with emerging intentions. At the affective layer, mild tension, curiosity, or heightened alertness may guide the revision process, prompting exploratory reformulation rather than mere error correction.

The cognitive layer becomes more prominent, drawing on memory, inference, and contextual integration to evaluate how revised segments fit within the whole. Revision thus operates not as a simple "post-processing" stage but as an enactive loop in which earlier outputs become new inputs. By re-entering already produced text as sensory evidence, the translator can recalibrate their behavioral routines and refine affective orientations. This cyclical process helps stabilize the unfolding translation, improves coherence, and strengthens the translator's evolving sense of how the text should sound and mean.

#### 3.4.3 The Flow State

During the Flow state, the translator operates within a sufficiently stabilized coupling configuration that perception, action, and ongoing translation can proceed with relatively little need for deliberate reconfiguration. The translator does not cease to perceive, evaluate, or infer; rather, these processes become tightly coordinated with routinized sensorimotor activity. Previously established expectations about the source text, target formulation, and task environment allow incoming information to be integrated into an ongoing action trajectory without repeatedly interrupting it to establish a new course of action. Flow is therefore not simply a state of low cognitive effort, but a state of stable organism–environment coupling in which the current action policy remains viable.

Flow is consequently a coordinated Affective–Behavioral–Cognitive configuration: affective dynamics sustain the perceived relevance and viability of the current course of action; behavioral dynamics enact the established sensorimotor contingencies; and inferential dynamics operate largely through anticipatory, action-guiding expectations rather than requiring continuous explicit deliberation. The resulting coordination allows the translator to maintain a coherent translation trajectory while minimizing the need for extensive reorientation.

## 4. The Translating System

We model the ABC framework in terms of Active Inference (AIF, [31, 32]). AIF offers a powerful framework for understanding how biological agents regulate their sensorimotor interactions with the environment. It is a prominent idea in computational neuroscience that proposes the brain performs approximate Bayesian inference [31, 39], constantly generating predictions about the causes of sensory stimuli and updating these predictions by minimizing "surprise" (prediction error) or "uncertainty" (expected free energy).

Ramstead et al. [23] provide an "enactive interpretation" in which the predictive brain is not a representational device but an action-oriented engagement machine [6,22] that selects frugal, action-based routines to reduce demands on neural processing and facilitate adaptive success. Action, in this view, is a generative process that changes the environment and therefore a precondition and enabler of perception. However, action is also steered by an internal "generative model". As Ramstead et al. [23] clarify, such internal "generative models are more about the control and regulation of action than they are about figuring out what is 'out there' … They enable survival, rather than tracking truth. They model the acting organism, and are used by living systems to modulate their behaviour." Agents maintain viable coupling by constantly minimizing prediction errors within a temporally thick horizon where past, present, and anticipated future co-inform each other. In translation, this temporal thickness allows coherence and flow between source comprehension and target production.

### *4.1. Temporal Thickness*

The notion of "temporal thickness"[2] is crucial within AIF and enables agents to engage in "proactive, purposeful inference about their own future" [22]. Friston [31] refers to temporal thickness as *counterfactual depth*, the ability to simulate counterfactual possibilities, and likely consequences of action by integrating information across different timescales. In the context of translation, temporal thickness is evident when translators coordinate short-term resources, such as the immediate co-text of a word or phrase translations with syntactic integration, or long-term considerations, including cultural and conceptual mismatches between the source text (ST) and target text (TT). This temporal nested structure is mirrored in layers of the ABC architecture which can be described as a Translation Behavioral Style Space [36].

Translators dynamically coordinate enlanguaged affordances concurrently on multiple levels of processing, actively authoring fields of sense-making through sensorimotor loops. The cognitive effort required to navigate these multi-dimensional pathways is directly visible in gazing and typing activities. As we shall see, these bodily movements offer empirical support for translation as a temporally thick, participatory engagement, that enact linguistic values in situ.

### *4.2. Behavioral Correlates of Cognitive Processes*

Evidence of cognitive-inferential processing in translation process data appears as dynamic reconfigurations of the translator's coupling with the environment. Such reorganizations may involve consulting external resources (dictionaries, parallel texts, web searches) alongside exploratory gaze behavior and extended keystroke pauses, indicating information gathering. Reorganization patterns may also coincide with increased keystroke variability, dispersed visual attention, or complex revision sequences before stabilizing into new patterns of translation production. Higher-order metacognitive control manifests as anticipatory (re)organization of translator–environment relations, opening new possibilities for action, for example:

- strategic pauses at meaningful boundaries rather than mere difficulty points
- proactive resource use, where gaze shifts to reference materials precede emerging translation problems.

The eye–mind hypothesis provides an empirical point of contact between observable gaze behavior and the organization of translation activity. Just and Carpenter's eye-mind hypothesis [37] formulation assumes a close temporal relation between fixation and the processing of the fixated material, an assumption that has subsequently been widely employed in translation-process research. The DEUTS thesis stipulates that recurrent sensorimotor interaction of the translator with her environment become dynamically coupled and exert a reciprocal causal influence as external tools become constitutive in the translation process. When gaze, source-text uptake, tool interaction, and subsequent translation actions form a temporally integrated feedback loop on which the ongoing activity depends, the coupled bodily and environmental processes can be treated as components of the currently extended translating system. On this view, the boundary of the translating mind can expand when a resource such as a terminology tool becomes constitutively integrated into the recurrent gaze–action–environment loop, and contract again when that resource ceases to participate in the ongoing organization.

[2] https://loc.closertotruth.com/theory/friston-s-free-energy-principle-and-active-inference

### *4.3 Constitutive Coupling and Markov Blankets*

This fluctuating coupling architecture provides a basis for the temporal dynamics of the translating system. A coupling configuration specifies how translation is organized at a particular point in time: which parts of the affordance landscape are being taken up, how they are coordinated, and which patterns of interaction sustain the ongoing translation trajectory. The landscape of enlanguaged affordance is not simply a collection of external resources but a relational field of possibilities for perceiving, interpreting, evaluating, and acting, shaped by the material, linguistic, technological, and social environment.

An environmental resource can become constitutively integrated into translating activity when the affordance-mediated coupling that involves that resource becomes sufficiently recurrent, reciprocal, and organizationally indispensable to the ongoing trajectory. Constitutivity therefore does not require mutual physical alteration of the components involved. An ST, for example, may repeatedly reorganize the translator's gaze, attention, segmentation, rereading, and textual actions, while the translator's activity in turn determines how particular parts of the ST are accessed, selected, revisited, and acted upon. Conversely, a change in coupling configuration does not by itself imply a change in the constitution of the translating mind. Configurations may fluctuate rapidly while the underlying organization of the translating system (the mind) remains relatively stable.

The notion of constitutive coupling provides the basis for specifying the dynamical system to which a Markov blanket (MB) can be assigned. Formally, an MB is a statistical-dynamical boundary that separates a specified system from the remainder of its environment through the conditional dependencies linking internal, sensory, and active states. The theoretical identification of constitutive organization precedes the specification of the relevant system and, consequently, its MB. In an extended translating system, bodily and technological components may participate in the system's internal dynamics when they are recurrently integrated into its organization, while the sensory and active states through which that system interacts with the remaining environment constitute its MB. The MB thus characterizes the sensory-active interface of the currently constituted translating system, whereas coupling configurations describe how that system is organized in relation to the affordance landscape at a particular moment. A translator may shift gaze from the ST to a dictionary, terminology database, or the emerging TT, thereby changing the current coupling configuration, without changing the boundary of the constituted system. Similarly, a broader engagement with available contextual or technological affordances does not automatically entail an expanded MB. A change in MB becomes relevant only when a sustained reorganization of affordance-mediated coupling changes the constitution of the dynamical system itself.

From this perspective, an MB can be understood as a relatively slower-changing organizational boundary in relation to the faster fluctuations of coupling configuration. It should not be regarded as a fixed spatial enclosure around the translator, nor as a boundary that changes whenever attention shifts or a new resource is consulted. Rather, an MB tracks the statistical-dynamical boundary of the system once a particular organization of constitutive coupling has been established. This distinction is important for avoiding a form of cognitive bloat (see below) in which every causal interaction with the environment is treated as part of the translating mind.

MBs can be, and in biological systems typically are [35], hierarchically nested, with subsystems maintaining a degree of autonomy while participating in larger-scale dynamics. The ABC architecture similarly proposes a hierarchically organized system [36], in which affective, behavioral, and cognitive processes are dynamically coupled across multiple temporal scales. Such nesting allows relatively stable organizational patterns to coexist with faster fluctuations in local coupling. It also provides a formal framework for considering how activity at one temporal scale constrains and is constrained by activity at others. The ABC levels describe functional dimensions of translating activity, whereas MBs describe statistical-dynamical boundaries of specified systems.

The present paper does not specify or empirically demonstrate the conditional-independence relations required to establish a MB in the strict formal sense. The notion is therefore used here as a theoretical and formalizable characterization of system boundaries, rather than as an empirical claim that the translating system has been shown to possess a particular MB. A full formal treatment would require specifying the relevant state variables, causal dependencies, and conditional-independence structure and demonstrating that these satisfy the requirements of a MB.

### *4.4 HORF States and Constitutive Integration*

HORF states characterize the temporally extended affective organization of translating activity within an affordance landscape. A single HORF state may encompass a sequence of changing coupling configurations as the translator's engagement with the environment unfolds. Orientation may involve successive configurations through which the translator explores different source-text, contextual, or technological affordances. Hesitation may encompass repeated reconfigurations involving intensified monitoring, comparison, and the exploration of alternative affordances. Revision may involve alternating configurations of engagement with the emerging and previously produced target text. Flow may encompass a succession of relatively stable configurations through which an established trajectory is enacted. Thus, a change in coupling configuration does not necessarily entail a change in HORF state: a translator can remain in the same affective mode while repeatedly reorganizing how different affordances are taken up.

HORF states therefore do not designate isolated internal emotional states that are subsequently expressed in behavior. They characterize temporally extended affective modes of engagement through which the affordance landscape is explored, organized, and reorganized. Affect is neither reducible to processing difficulty nor to a particular level of precision weighting. Rather, affective orientation contributes to relevance realization by influencing which affordances are approached, explored, monitored, or re-evaluated and how strongly they guide ongoing activity. These affectively organized patterns of engagement can, in turn, reorganize the coupling configuration and, when a new organization of affordance-mediated activity becomes sufficiently recurrent and constitutive, contribute to a reconfiguration of the translating system.

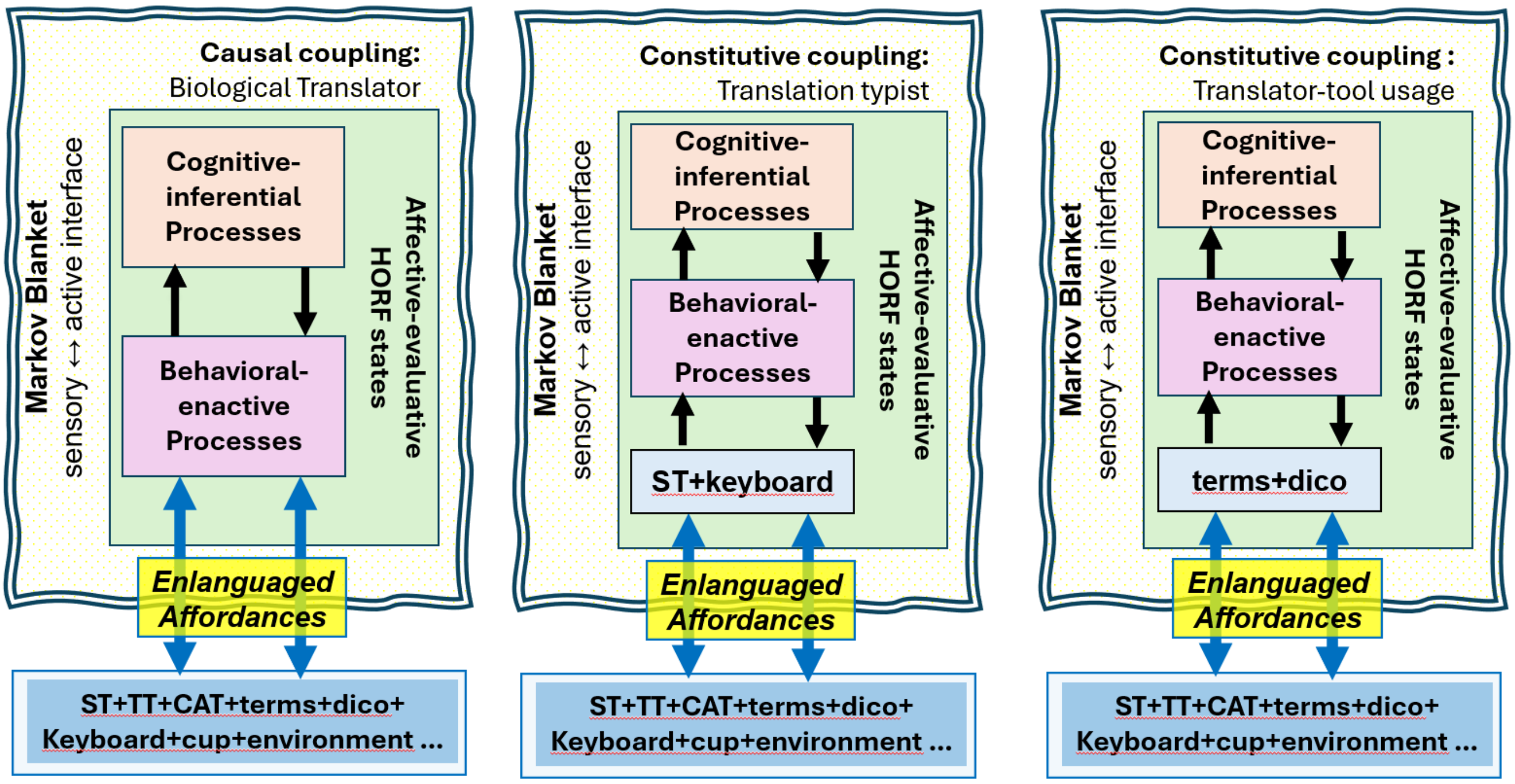

**Figure 2.** The three scenarios illustrate how the translating system (i.e., the mind) can dynamically reconfigure its coupling with the environment. Different material components may become temporarily constitutive through recurrent interaction mediated by enlanguaged affordances.

The left panel depicts the biological translator, organized through dynamically coupled behavioral-enactive and cognitive-inferential processes, with affective-evaluative dynamics expressed through HORF states. These processes are causally coupled to an external environment comprising the ST, TT, computer-assisted translation tools (CAT), terminology resources, dictionaries, keyboards, and other material resources through enlanguaged affordances. The sensory-active interface constitutes the MB of the currently specified translating system.

In the middle panel, the translator's activity is organized through coupling with the ST and keyboard. Enlanguaged affordances are constitutively integrated into the translating system through the recurrent organization of the translator's activity. It is the affordance-mediated coupling with material components of the environment that becomes constitutively organized within the translating activity.

In the right panel, a translation problem leads the translator to engage a dictionary and term base. Their enlanguaged affordances become part of the current coupling configuration as the translator–gaze–tool–action interaction becomes recurrently organized around the translation problem. The tools may thereby become constitutively integrated into the translating system for the duration of this activity. Once the relevant information has been incorporated into the evolving target text, the translator closes the tool and returns to the previous coupling configuration.

The figure thus illustrates that the material components of the translation environment can become temporarily constitutive through recurrent coupling mediated by enlanguaged affordances. The blue vertical arrows indicate the permeability of the system's MB with environmental states through the landscape of enlanguaged affordances.

HORF states do not entail a change in the MB. The same constitutively organized translating system can pass through different HORF states while retaining the same system boundary. Conversely, a single HORF state can encompass multiple coupling configurations without requiring any change in the MB. A change in the MB becomes relevant only when the ongoing reorganization of affordance-mediated coupling changes the constitution of the dynamical system itself. In this sense, the MB should not be understood as tracking moment-to-moment changes in attention, task focus, or affordance selection.

Thus, HORF states characterize the affective organization of temporally extended translating activity; coupling configurations characterize the changing organization of engagement with the affordance landscape; and the MB characterizes the sensory-active interface of the currently constituted dynamical system.

$$
\begin{aligned}
C_t &::\text{current coupling configuration}\\
HORF_t &::\text{affective mode in which } C_t \text{ is enacted}\\
MB_t &::\text{sensory-active interface of the currently constituted system}
\end{aligned}
$$

A change in coupling configuration does not by itself imply a change in HORF state or in the MB, and a change in HORF state does not by itself imply a change in the MB.

*4.5 The Translators' Markov Blanket*

We separate the functional, information-processing boundary that separates the constitutive states of the translation process from the causal background factors of the environment. When a translator couples deeply with a translation tool, a new, larger MB emerges and wraps around both of them. This MB is a dynamic, permeable boundary that shifts in real-time which flexes and redraws based on fluctuating attentional focus, expanding to incorporate more external context during analytical problem-solving and tightening during intuitive flow. External states are states outside the system's current constitutive organization, but they remain causally coupled to it through the MB and may become incorporated into the constitutive loop through perception and action. Hence, an MB can move, that is, the boundary of what is constitutive of the translation process can expand or contract. Figure 2 shows three coupling configurations with different MBs that conceptually separate the translator's internal ABC states from the external/environmental states.

1. When translation is flowing smoothly, the constitutive process requires very little external input to maintain its trajectory. The translation loop is highly localized and self-contained. The MB contracts tightly around a small, automated loop, the text on the page is immediately integrated into the constitutive process, the visual input (gaze) is instantly ingested into the constitutive execution loop, translating directly into motor output (keystrokes). The causal external world is kept at a distance because the internal predictive processing is highly efficient and self-sufficient.

2. When a translation breaks down, e.g., a complex syntactic inversion or an ambiguous metaphor occurs, the existing internal states can no longer resolve the uncertainty. To minimize free energy, the system must expand its boundaries to pull external, previously causal elements into the constitutive process. The MB expands outward, so that what was previously just a "causal environmental factor" (e.g., the next paragraph, a parallel reference text, a dictionary entry, a terminology tool) is actively drawn inside the MB, becoming a constitutive part of the active problem-solving system. The temporal gap between gaze and keystroke stretches as the system is expanding its informational bandwidth. The gaze may jump across the page to absorb these external causal factors. Keystrokes pause because the constitutive loop is busy restructuring itself to integrate this newly internalized information before it can resume motor execution.

### 4.6 The Cognitive Bloat Problem.

The cognitive-bloat problem [38] arises if every object that causally interacts with a translator—source text, keyboard, dictionary, terminology database, screen, or online resource—is thereby counted as part of the translating mind. The extended-mind hypothesis proposes that external resources can sometimes function as physical extensions of cognition, but this does not entail that every causal interaction is constitutive. Suppose there is a coffee cup next to the translator which she uses. It is part of the physical environment, but as it does not participate in the organization of the translating activity through a relevant affordance, there is no reason to include it in the translating mind. Several complementary criteria can constrain such extension.

1. **Trust and glue criteria** [1,40]. The *trust* criterion asks whether the agent reliably delegates part of an ongoing cognitive activity to the external resource, treating its operation as dependable enough to incorporate into action. The *glue* criterion asks whether the resource is functionally integrated into the broader activity such that the agent's behavior is organized around its continued availability, rather than merely consulting it occasionally. Together, these criteria help distinguish an integrated component of an extended system from an external resource that is simply used.
2. **Dependence and indispensability.** A resource is a stronger candidate for constitutive status when its recurrent participation is necessary for maintaining and regulating the characteristic activity. Removing or substantially altering the resource should therefore produce a systematic reorganization of the activity, rather than simply depriving the translator of one optional source of information. A tool that can be removed without substantially reorganizing the ongoing activity is more plausibly merely causal; one that has become transparently integrated into a recurrent sensorimotor loop is a stronger candidate for constitutive status.

3. **DEUTS and dynamical integration.** Under DEUTS, an environmental resource becomes constitutive when it participates in a recurrent and tightly integrated organization of the activity, rather than merely exerting an external causal influence. Constitutive coupling involves reciprocal organization of the coupling: the resource constrains and supports the ongoing activity, while the agent's activity continuously reorganizes its relation to the resource.
4. **Markov-blanket criterion.** Kirchhoff and Kiverstein [2] proposed using Markov blankets to provide principled criteria for demarcating the boundaries of cognitive systems. For a candidate system with internal states $I$, external states $E$, and blanket states $B$, the conditional-independence relation:
$$p(I, E \mid B) = p(I \mid B)p(E \mid B)$$
provides a formal constraint on the system boundary. The MB therefore describes the interface of a system once that system has been specified. However, it does not, by itself, determine which environmental resources are constitutive.

Taken together, these criteria distinguish causal from constitutive coupling without requiring a rigid boundary between organism and environment. They also provide a principled basis for determining when a change in coupling configuration warrants treating an environmental resource as part of a newly constituted system—and hence when a change in the corresponding MB is justified.

## 5. Head-starter and Context Planner

The illustration of a translation for an English ST sentence into Japanese is shown in Figures 3 and 4. Both Figures show the same English ST sentence which consists of four chunks, shown in Table 1.

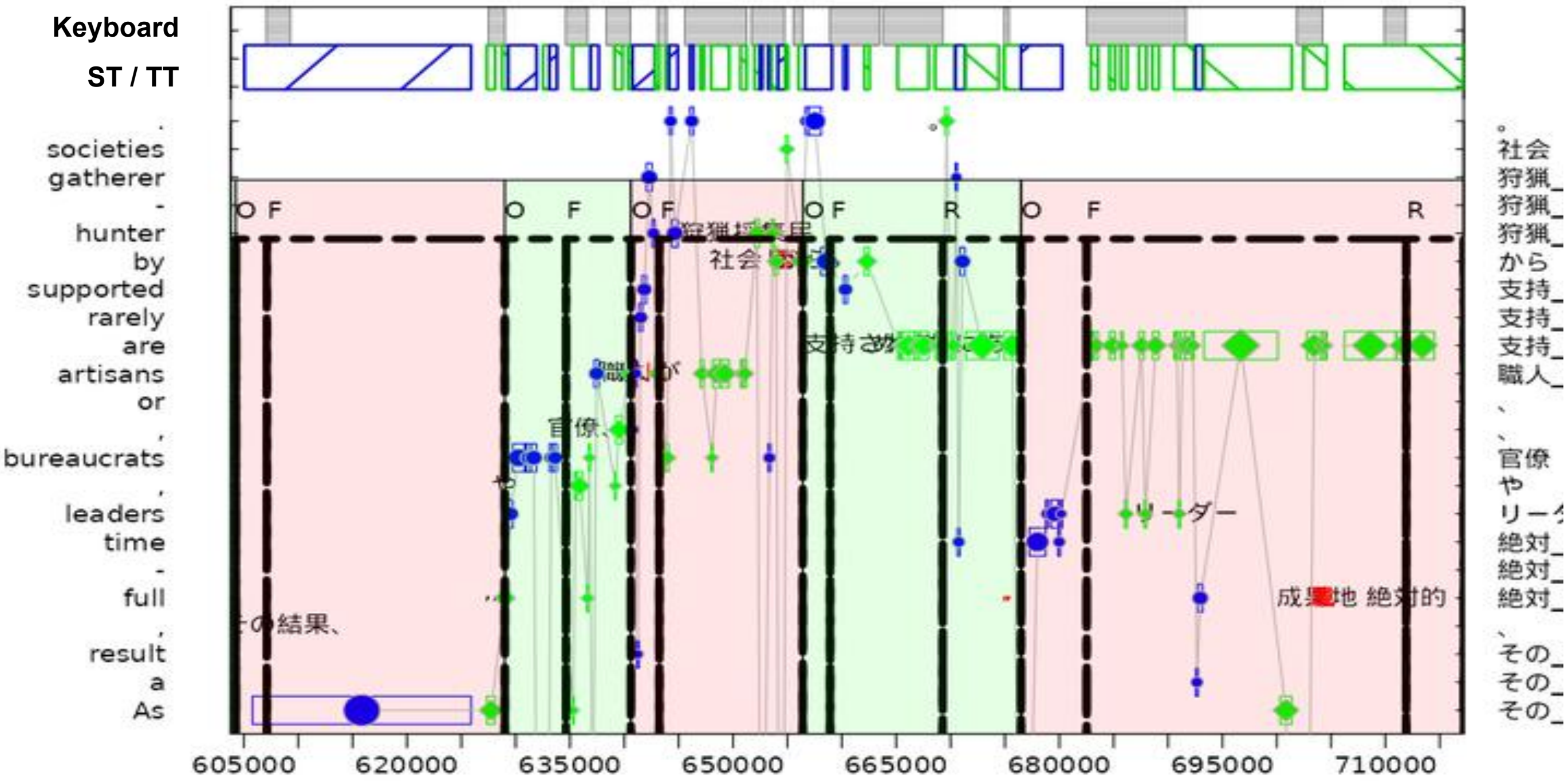

**Figure 3.** The Progression Graph shows a segment of approx. 105 seconds for translator (i) producing a linear translation from English to Japanese. The x-axis denotes translation time in ms (*sec.* 605 – 710); the left y-axis lists source words. The plot visualizes gazing and keystroke behavior. Blue dots mark fixations on the ST; green dots mark fixations on the TT; Japanese characters are the keystrokes. The top black striped bars indicate segments of coherent typing, i.e., keyboard coupling; the blue and green boxes show coupling with ST and TT respectively. The plot also shows HORF states (Hesitation, Orientation, Revision, Flow) and their grouping into policies. The graph shows that the translation was produced in five policies: three OF followed by two OFR, marked alternating in green and red striped boxes.

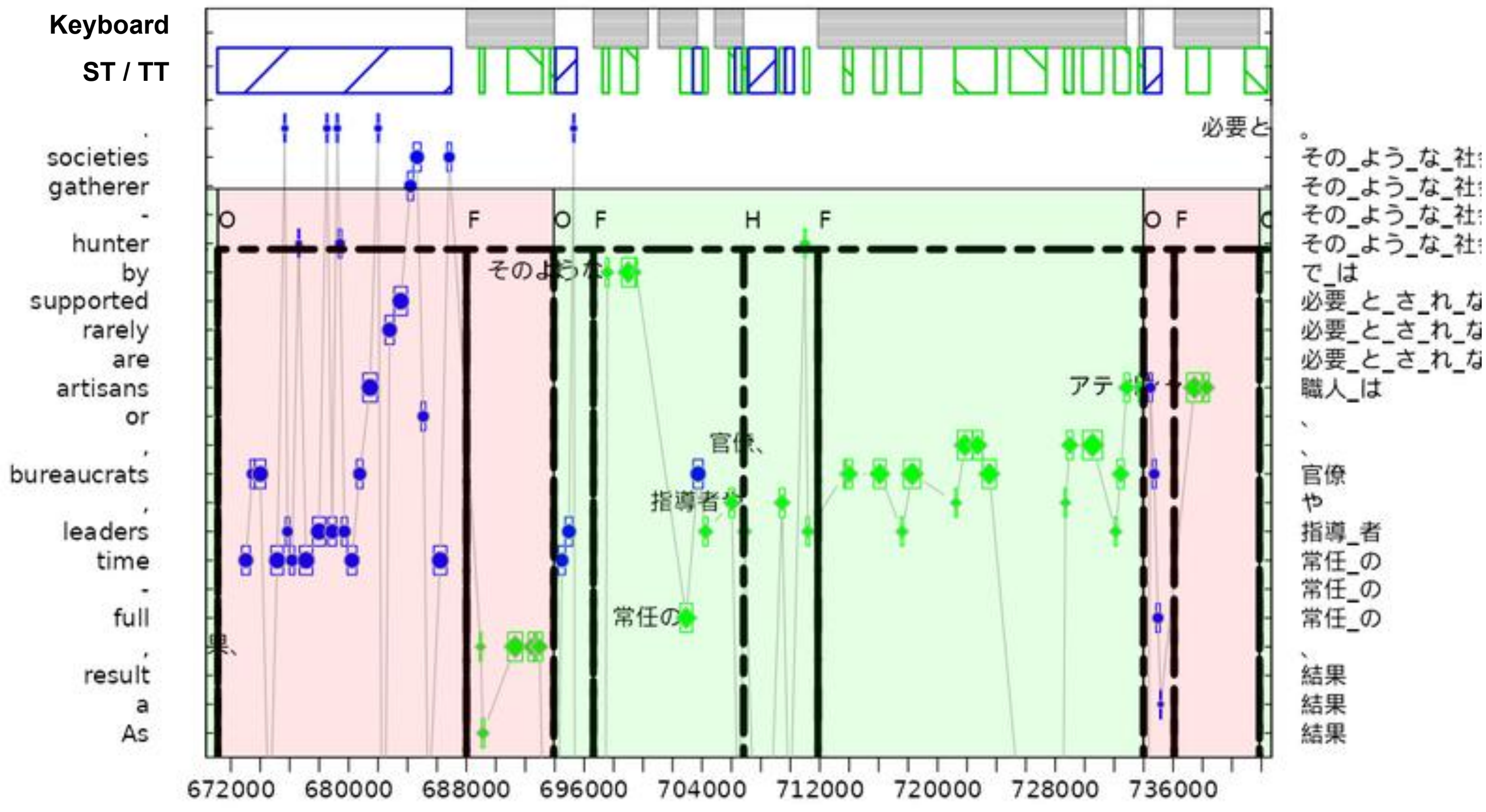

**Figure 4.** The Progression Graph shows a segment of approx. 64 sec. for translator (ii) producing a reversed translation for the same ST that was achieved by means of three policies: OF-OFHF-OF.

**Table 1.** English chunks and their Japanese translations.

| # | English Chunk | Translation (i) | Translation (ii) |
|---|---|---|---|
| ① | As a result, | その結果、 | 結果、 |
| ② | full-time leaders, bureau-crats, or artisans | 絶対的リーダーや官僚、職人が | 常任の指導者や官僚、職人は |
| ③ | are rarely supported | 支持されることは、めったにありませんでした | 必要とされない |
| ④ | by hunter-gatherer societies | 狩猟採集民族社会から | そのような社会では |

The English sentence shown in Table 1 has been translated into Japanese in a linear order ①, ②, ④, ③ by translator (i) in Figure 3, with a high degree ST similarity, moving the translation of the verbal phrase ③ to the final position, in line with Japanese syntactic constraints. Translator (ii), by contrast, has chosen a reverse-order translation ①, ④, ②, ③ which brings to the front the prepositional phrase ④ in the Japanese translation. While both versions are valid Japanese translations, they have different behavioral implications.

### *5.1 Translation Styles*

Translator (i) can be characterized as 'head starter' [42], choosing to read and type the translations chunk by chunk as they go along in the text. Epistemic states of orientation are short, with a preference for pragmatic flow states. For instance, until around time=628,000, the translator only fixates chunk ① while typing its translation, indicating minimal lookahead.

On the other hand, the translator (ii) begins with a much longer orientation phase, where the entire sentence is scanned before any words are typed. This translation strategy can be characterized as "large context planner" [42]. Note that the successive orientation states are relatively short.

The two different translation styles result in different policy cycles. For translator (ii), a long initial state of orientation indicates the need for epistemic value, whereas the head starter engages immediately in pragmatic translation typing. While the head starter produces more subsequent revisions (marked **R**, in the Figures), the more cautious large context planner has more hesitation (**H**, e.g. Figure 4 around time=710,000).

### *5.2 Coupling Configurations and Translation Styles*

The two translation trajectories can be interpreted as different patterns of coupling between the translator and the material and linguistic environment. From the perspective developed above, the MB characterizes the sensory-active interface of a currently constituted translating system, i.e., the translating mind. The translator's coupling configuration can change substantially while the MB remains unchanged.

For the head-starter, coupling is initially organized around a relatively local portion of the translation environment. Translator (i) engages primarily with the currently processed ST segment, the emerging TT, and the immediate sensorimotor resources through which the translation is enacted. The enlanguaged affordances of these material components—reading, interpreting, producing, inspecting, and revising—are repeatedly taken up in successive cycles of perception and action. Rather than sustaining coordination with a broader portion of the ST before acting, the translator progressively resolves uncertainty through these local cycles of engagement.

The resulting trajectory can therefore be characterized as a sequence of relatively local coupling configurations. Each typing action modifies the material state of the emerging TT, and this modified artifact subsequently provides new perceptual and affordances for the next step. The reduction of uncertainty thus occurs primarily through enactment and successive reorganization of the local coupling configuration. However, this does not require the MB itself to change with every new action: changes in gaze, attention, or task focus constitute changes in coupling configuration, not necessarily changes in the boundary of the translating system.

The large-context planner, by contrast, engages in a broader coupling configuration, probably activating a larger landscape of enlanguaged affordances before target-text production is committed. Uncertainty is consequently reduced through expanded exploration and coordination within the current coupling configuration, rather than primarily through rapid successive acts of local enactment.

The engagement of an external resource does not automatically make that resource part of the translating mind. A dictionary, terminology database, or translation memory becomes constitutively integrated only insofar as its affordances participate in a recurrent organization of the translating activity. For example, a translator–gaze–dictionary–action loop may temporarily incorporate the dictionary into the ongoing organization of translation when consulting, evaluating, and acting become recurrently interdependent. The relevant issue is therefore not whether the resource is physically external, but whether its affordances have become constitutively integrated into the current translating system.

In both trajectories, the organization of coupling can fluctuate continuously as the translator engages with different material and linguistic resources. The difference lies primarily in the temporal scale, breadth, and persistence of coupling: the head-starter tends to stabilize relatively local configurations and move rapidly from one configuration to the next. The large-context planner tends to sustain a broader configuration for longer before committing to a target-text action.

*5.3 Coupling Configurations and HORF states*

HORF states specify the affective attitude in which the coupling is enacted. Different translation trajectories pass through Orientation, Flow, Hesitation, and Revision states, with potentially different temporal organizations of those states.

The head-starter is expected to spend relatively more time in local Flow-like coupling, in which the current source-text segment and emerging target text provide sufficient structure for continued action. The transition from one chunk to the next can occur without substantial expansion of the coupling configuration. When an unexpected lexical, syntactic, or contextual problem is encountered, the translator may enter Hesitation, temporarily disrupting the established coupling. Resolution can then occur through a brief expansion of the coupling configuration.

The large-context planner, in contrast, begins with a more extended orientation state. Orientation here is not merely an internal preparation stage but an active expansion and calibration: the translator samples a broader portion of the source text and establishes which elements are likely to be relevant for subsequent action. Once this broader configuration becomes sufficiently stable, the translator can enter Flow and produce target text, probably with fewer local disruptions. The longer initial orientation therefore trades a greater investment in environmental exploration for potentially greater stability during subsequent enactment.

States of Hesitation and Revision mark moments at which the existing coupling configuration no longer adequately supports the current trajectory. Hesitation may correspond to a temporary increase in uncertainty that prompts the translator to explore additional affordances or recruit previously inactive resources. Revision may involve a more substantial reorganization, in which the translator changes the relationship between the source text, the emerging target text, and the resources currently included in the effective coupling. In both cases, a change in HORF can therefore accompany—and potentially contribute to—a change in the effective MB.

The relationship is consequently reciprocal: on the one hand, $MB_t \rightarrow HORF_{t+1}$ as the current coupling configuration constrains the possibilities available for action and influences their affective significance. On the other hand: $HORF_t \rightarrow MB_{t+1}$ because the translator's current affective state influences which environmental resources may become relevant, which affordances are explored, and whether the current coupling is maintained or reorganized.

The two trajectories illustrate two different ways in which uncertainty, coupling, and affective regulation can become temporally organized. The head-starter primarily resolves uncertainty by moving rapidly through successive local coupling configurations, whereas the large-context planner temporarily expands and stabilizes its coupling configuration before committing to action. In both cases, HORF states describe the affective modes through which these coupling configurations are enacted.

*5.4 Relation to ABC*

The difference between the trajectories of the head-starter and large-context panner emerges from the interaction of all three ABC layers:

- Affective-evaluative dynamics influence the relevance and urgency of available possibilities and help regulate whether the current coupling is maintained, explored, or reorganized.
- Behavioral-enacting dynamics enact the coupling through gaze, typing, cursor movements, resource consultation, etc.

- Cognitive-inferential dynamics reorganize the distribution over viable action trajectories when established coupling does not sufficiently constrain what to do next.

It illustrates a central claim of the ABC framework: translation is organized neither by an isolated internal cognitive model nor by environmental input alone, but by the recurrent coordination of affective-evaluative, behavioral-enacting, and cognitive-inferential dynamics across a fluctuating organism–environment boundary. Within this coupling, enlanguaged affordances provide historically sedimented possibilities for understanding, formulation, inference, and coordinated action that become available through the translator's engagement with the ST, TT, terminology, translation conventions, and other linguistic and socio-cultural resources. They are not treated as internally stored representations, but as socially stabilized possibilities for action and sense-making that can be selectively taken up, enacted, and transformed into situated activity. Their relevance is therefore dynamically realized rather than simply retrieved. The example illustrates how different temporal boundaries of the MBs, coupling configurations, HORF dynamics, and access to enlanguaged affordances can generate different patterns of translation behavior.

## 6 Enactive Inference

We present an enactive-inference model that illustrates two translation strategies: (i) linear translation, which proceeds incrementally through source-text chunks, and (ii) reverse-order translation, which involves greater reordering and broader exploration of the source-text affordance landscape before target-text production. As depicted in Figure 2, the model comprises Internal, External, Sensory, and Active states that form closed perception–action loops. These distinctions should not be interpreted as a division between an internal representational system and an external world. Rather, they specify the states and interactions used to formalize the current organization of the translating system and its coupling with the environment. Active states alter the translator's engagement with the environment—for example, through typing, gaze shifts, rereading, or resource consultation—and thereby modify subsequent sensory states and the affordances available for further action.

Following Mizowaki [43], the model is implemented as a partially observable Markov decision process (POMDP, [44, 45]) with two levels of organization: word-level and chunk-level. Within this formalism, probability distributions over states are not interpreted as representations of independently existing translation states. Rather, they characterize the currently viable action-relevant trajectories available to the translating system under uncertainty. The model therefore captures how alternative trajectories of engagement are differentially stabilized as translation unfolds. Inferential dynamics reorganize these distributions in response to ongoing sensorimotor coupling, while behavioral dynamics enact policies through typing, gaze shifts, resource consultation, rereading, and reformulation. Affective dynamics, represented through the HORF organization of translating activity, modulate the relevance and persistence of alternative trajectories and thereby contribute to the ongoing realization of relevance within the affordance landscape.

The two modeled translation strategies can consequently be understood not simply as different sequences of linguistic operations, but as different temporal organizations of coupling. Linear translation tends to stabilize relatively local coupling configurations and move incrementally from one source-text segment to the next. Reverse-order translation involves more sustained exploration and coordination of affordances across a broader temporal and textual context before a target-text action is enacted. In both cases, the translating system continually reconfigures its engagement with the affordance landscape while maintaining a relatively stable organizational boundary.

The formalism thus provides a way of modeling enacted uncertainty reduction without assuming that translation depends on the internal storage and manipulation of symbolic representations of source or target text. Internal states in the POMDP are formal variables in the model, not claims about an inner representational medium. The model instead describes how affectively organized, action-guiding dynamics select, stabilize, and reorganize trajectories of engagement as the translating system unfolds over time.

### *6.1. Simulation and Evaluation*

Here, we present a high-level description of the enactive inference process based on the ST and TT inputs from Table 1. We only discuss word order choices in the target language, putting aside for the moment the difficult problem of lexical selection. While lexical translation ambiguity plays an important role in policy selection, for the sake of simplicity we focus here on word-order. Thus, based on the four chunk translations from Table 1:

- ① その結果
- ② 絶対的リーダーや官僚、職人が
- ③ 支持されることは、めったにありませんでした
- ④ 狩猟採集民族社会から

we generated the six Japanese six translation trajectories (translation candidates: $TT_0$ to $TT_5$, in Table 2) by holding the vocabulary constant and manipulating only the chunk order. Here, $TT_0$ reproduces the linear translation (i), while $TT_3$ reproduces the reverse-order translation (ii), albeit with a different vocabulary. All together, these four chunks can be combined into six Japanese translations in different ways:

**Table 2.** Chunk orders for translations $TT_0$–$TT_5$. The source chunk order (for reference) is ①, ②, ③, ④. We consider commas ("、") to be one of the chunks as well.

| Candidate Translation | Positions of chunk translations in TT | | | | |
|---|---|---|---|---|---|
| | 1 | 2 | 3 | 4 | 5 |
| $TT_0$ | ① | 、 | ② | ④ | ③ |
| $TT_1$ | ② | ① | 、 | ④ | ③ |
| $TT_2$ | ① | ② | 、 | ④ | ③ |
| $TT_3$ | ① | 、 | ④ | ② | ③ |
| $TT_4$ | ① | ④ | 、 | ② | ③ |
| $TT_5$ | ④ | ① | 、 | ② | ③ |

As shown in Table 2, $TT_0$–$TT_2$ preserve an order relatively close to the source — phrase ④ appears in the latter half — whereas the position of phrase ② varies across candidates. $TT_3$–$TT_5$ implement substantial reordering: phrase ④ appears in the first half, again with candidate-specific positions. The verbal chunk ③ appears sentence final in all translations.

Because the chunk translations are identical across all candidates, the lexical entropy is zero for all translations. By contrast, the positional entropy is larger than zero for all chunks. Except for chunk ③, the word translations appear in different positions across translations from $TT_0$ to $TT_5$. In particular, the chunks ② and ④ exhibit large positional variability across candidate translations and therefore carry high positional information. In ABC terms, the Affective dynamics modulate how strongly entropy reductions influence subsequent policy selection at the Behavioral layer and the stabilization of action-relevant expectations at the inferential layer.

### *6.2 Translation Strategies*

Before the translator starts reading any ST words, the mapping between ST chunks and TT positions is highly uncertain, and thus the chunk-order entropy is high. Successively, as the translator reads chunk by chunk sequentially, the (posterior) entropy reduces because each new chunk narrows the distribution of possible TT orderings. When reading Chunks ① and ③ the model entropy drops quickly because their positions are relatively stable. Chunk ① translation can only appear in position 1 and 2 in the Japanese sentence, while chunk ③ translation appears only at position 5. When reading Chunks ② and ④ the entropy reductions happen more slowly, as they carry most of the "informational load" in the mapping task, these chunk translations can occur in all target position, except in position 5.

Conversely, when typing the chunk translations, the biggest entropy drop happens when placing ② and especially ④, since those chunks have the highest positional variability across the six candidate translations. Their placement will hence decrease most of the variability in the sentence. By contrast, placing chunk ① reduces the sentence entropy only slightly and typing chunk ③ does no reduce the word-order entropy at all, as there is no positional uncertainty related to the placement of this chunk.

### *6.3. Head-Starter vs. Large-Context Planner*

The head-starter and the large-context planner represent two contrasting ways of organizing translation under uncertainty. The head-starter resolves uncertainty primarily through action: target-text production begins soon after local source-text processing, and subsequent actions progressively stabilize the translation trajectory. For source languages such as Japanese, where important structural information may arrive late in the sentence, this incremental strategy can support rapid production and reduce the need to maintain extensive source-text context. Its disadvantage is that early commitments may have to be reorganized when later source-text information makes a different target-text structure more appropriate, potentially resulting in reordering and revision.

The large-context planner, by contrast, reduces uncertainty through more extended engagement with the affordance landscape before overt target-text action. By coordinating a broader source-text context, it can establish a more globally coherent target-text trajectory before committing to production. This may reduce subsequent reordering and revision, but at the cost of delayed production and greater demands on sustained attention and contextual maintenance.

These strategies can be understood as different temporal organizations of coupling. The head-starter rapidly reconfigures local coupling through successive perception–action cycles, whereas the large-context planner sustains broader coupling configurations before committing to action. The distinction therefore concerns where and when uncertainty is resolved within the translating trajectory rather than two fundamentally different cognitive architectures.

Within the ABC framework, behavioral dynamics enact policies through actions such as typing, reordering, gaze shifts, resource consultation, and revision. Inferential dynamics reorganize action-relevant anticipatory dispositions as new information or interactional constraints emerge. Affective dynamics modulate the relevance and significance of uncertainty, influencing whether the system maintains an established trajectory, explores alternatives, or expands its engagement with additional affordances. The formal probability distributions of the model consequently describe the system's current disposition toward alternative action trajectories rather than representations of a hidden "true" translation order.

Thus, the head-starter and large-context planner differ primarily in how uncertainty is temporally distributed across coupling: the former resolves it predominantly through incremental enactment, while the latter sustains broader contextual engagement before overt commitment.

## 7. Conclusion

Hutchins' [46] "Cognition in the Wild" can be considered a precursor of the "The Extended Mind" (EM) hypothesis which is, however, strongly rooted in classical cognitivism. Hutchins maintains that cognition is a form of computation which implies the "propagation of representational state[s] across representational media" [46]. These representational states are propagated across different internal and external media "by bringing them into coordination with one another." Some of the structure, he says, "is internal to the individuals and some is external." [46] The truthful re-representation of these structures across the different media is thereby based on a set of "axiomatic propositions and a set of rules", which "preserve the truth of the axioms." [46] External tools and artifacts enter this chain of mediated structures, which form a rich network of mutual computational and representational dependencies, in which "each tool creates the environment for the others" [34] With the extension of the mind into the 'Wild', humans "create their cognitive powers by creating the environments in which they exercise those powers" thereby dissolving the "boundaries of the skin." [46]

Kirchhoff & Kiverstein [2] analyze successive extensions of the EM hypothesis. They make out three "waves" of thinking about the extended mind. According to them, the first wave EM framework [1] stipulates the functional similarities between internal and external tools, such when a notebook takes over or replaces internal biological memory. Second-wave EM "stresses different but complementary functional properties" [2] of the brain and external resources, in which "non-biological scaffolding" augments the brain's biological modes of processing. These external 'scaffoldings' are "transformatory of our cognitive capacities", they produce "new cognitive powers … because of the different functional properties" [2].

The third wave conceptualizes the mind as "diachronically constituted." That is, the mind is not fixed or static, but negotiable and fragile. It is constituted through the unfolding history of engagements with cultural practices and conditioned on how the agent has been historically embedded in practices. The diachronic perspective rejects "the idea of complementarity" [2] which is part of the second EM wave, and which stipulates a "fixed-properties view" of internal and external elements. Kirchhoff & Kiverstein [2] endorse the notion of "active externalism" in which elements of the environment play "an active role in driving cognitive processes." Cognition is here a pattern of active coupling in which the agent's movements, attention, affect, and environment jointly bring forth cognitive states.

While this echoes Hutchins's [46] point that cognition is the propagation of activity across various internal and external 'media', the third wave EM rejects representationalism all together. Third wave EM protagonists refute the notion of "mental representation" as unnecessary or even misleading: meaning emerges from temporally thick, skillful engagement, not from internal symbolic encoding. The mind *just is* a set of dynamic patterns across brain, body, and world. Ramstead et al. [41] build on this third wave EM, re-formulating key-components of Active Inference [32] in which the "generative model" (i.e., the agent's internal structure) is not a representational device but a control structure that allows to optimize interaction by reducing the free energy [34]. In this "enactive inference" framework, the brain does not represent, but it controls and regulates mental processes, i.e., processes that stretch into the environment.

Instead of modelling translation as "inner representation + output," we suggest, in line with the third EM wave seeing translation as an ongoing practice in which text, translator, tools, and sociocultural norms form a single extended loop. While cognition is the ongoing extended process, the mind is the emergent, affectively and normatively structured pattern of that process, not a separate substance, but the system's self-organized identity. Cognition is the mechanism, while the mind is the emergent organization of that mechanism.

The ABC model frames translation not as an inner symbolic process aided by tools, but as an extended, affectively attuned practice in which brain, body, tools, and cultural norms co-realize meaning in real time. In this view, the translator's embodied and affective attunement to their work environment (software interfaces, source texts, glossaries, client expectations, time pressure, ergonomics, etc.) is part of the cognitive–affective loop, not just a backdrop.

We discuss an example in which we model translation as Predictive Processing/Enactive Inference: each translation step can be seen as the minimization of prediction error across multiple layers of the ABC system. At the affective layer, precision weighting determines how strongly certain cues (e.g., translator habits, stylistic conventions) modulate decision-making. At the behavioral layer, gaze movements, typing, and pauses reflect the agent's ongoing adjustments to minimize discrepancies between expected and actual task progress. At the cognitive layer, higher-level inferences about semantic fit, coherence, and adequacy are continuously revised as new linguistic material is processed.

Affect is thereby conceptualized not as discrete emotional episodes but as encompassing the physiological and psychological underpinnings of emotion, including stress, arousal, and embodied responses to linguistic stimuli. In this view, affect shapes the allocation of attentional resources, the efficiency of decision-making, and the quality of translational output, while cognitive activity is inseparable from affective influences. This is largely in line with recent discussions of emotions in cognitive translation and interpreting studies (e.g., [47])

The ABC framework is also compatible with empirical findings in CTIS, e.g., that affect can both facilitate and hinder translational performance. In this respect, studies of affective congruence suggest that alignment between translators' own emotional states and the emotional valence of the source text, fosters fluency, creativity, and narrative engagement, while incongruence can impede comprehension and slow response times (e.g., [48, 49, 50]). In the ABC view, this results from the interaction among the three layers, where signals from lower levels - through action-perception loops and engagement with the environment - incur levels of surprise and uncertainty at the affective layer, modulating the actions taken by the translator. This modulation, additionally, is subject to individual differences, task demands, and L1/L2 directionality (e.g., [51]).

The ABC framework posits that mental processes encompass not only the source text, speakers, or multimodal stimuli, but also the emotions of others, which may evoke empathy or stress. The affective states can either mobilize cognitive resources (facilitating performance) or overwhelm them, depending on intensity. Other than the textual or multimodal materials, affect can also emerge through interpersonal and contextual processes, and this also means that translators and interpreters can be susceptible to "emotional contagion" -- the emotions embedded in source texts or conveyed by speakers are internalized and mirrored physiologically. Interpreting in traumatic or highly emotional contexts can induce vicarious stress or even secondary trauma, while more subtle contexts, such as politically charged discourse, can bias decision-making in ways aligned with translators' own ideological stance. All these aspects can be described as the consequences of expected free energy minimization which in turn, is strongly influenced by affective precision modulation. Precision modulation determines how strongly affective states weigh the relative influence of the epistemic and pragmatic drive, thereby constraining the minimization of expected free energy. For example, when affect is calm and confident, greater precision is assigned to predictions. This increases the expected pragmatic value through preferences at the layers below and reduces exploratory behavior. On the other hand, a cautious, anxious, or fatigued affective state increases uncertainty, thus increasing the epistemic component of expected free energy, driving information-seeking behaviour.

In summary, our ABC framework treats affect not as an incidental by product but as a constitutive element of how translators process linguistic input, interact with the external environment, regulate attention, construct meaning, and produce the target text.

The ABC framework, like Active Inference (AIF), should not be regarded as a narrow theory but more as a principled framework for understanding the dynamics of translation. Still, for it to have scientific credibility, it must articulate conditions under which its claims could be challenged. These falsifiability criteria include:

- **Precision-modulation claims**. The framework posits that the Affective layer tunes the precision with which the Behavioral and Cognitive layers weigh alternative mappings. This could be falsified if experimental evidence showed that translators' affective states (e.g., stress, confidence, fatigue) have no measurable effect on decision-making speed or variability in translation choices.
- **Entropy-reduction dynamics**. The framework assumes that translation unfolds as a progressive reduction of uncertainty (model entropy) coupled with fluctuating process entropy during action. It would be challenged if empirical data showed no systematic entropy reduction as translators advance through text chunks.
- **Layer interaction.** If Cognitive (affective-evaluative-updating), Behavioral (policy selection), and Affective (precision modulation) layers could be empirically shown to operate fully independently, with no measurable influence across layers, the core principle of integrative coupling would be falsified.

It must be acknowledged that the present study demonstrates only the basics of an ABC framework. While we provide an example that illustrates how uncertainty reduction and precision-modulated action can be modeled, it does not capture the full integration of Affective, Behavioral, and Cognitive dynamics in real-world translation.

## Appendix A. Active Inference Quantities

Active inference and predictive processing rely on a family of related but distinct quantities. Because these are sometimes used interchangeably in informal discussion, we define each one here, indicate what it measures, and state how it is used in this article. Throughout, $o$ denotes a sensory outcome (observation), $s$ a hidden (external or latent) state, $P$ the agent's generative model, and $q$ an approximate posterior belief. $E_q[\cdot]$ denotes expectation under $q$; $H[\cdot]$ denotes Shannon entropy; and $D_{KL}[\cdot \| \cdot]$ denotes the Kullback–Leibler divergence. The two composite objectives, variational free energy and expected free energy, are built from the simpler quantities.

**Table A1.** Glossary of active-inference quantities used in this article.

| Expression | Description |
|---|---|
| $\text{surprise}(o) = -\ln P(o)$<br>$H[P(x)] = -\sum_x P(x) \ln P(x) = E_P[-\ln P(x)]$ | **Surprise** (surprisal): the negative log-probability of an outcome.<br>**Entropy**: the expected (average) surprise of a distribution (its overall uncertainty). |
| $\varepsilon = o - g(\mu)$<br>precision-weighted: $\Pi\, \varepsilon$ | **Prediction error**: the mismatch between an observation $o$ and the value predicted from current beliefs $\mu$ (through the generative mapping $g$), scaled by the precision $\Pi$. Minimizing precision-weighted prediction error is how variational free energy is minimized in predictive-coding schemes. |
| $\Pi = \Sigma^{-1}$<br>uncertainty $\propto H[q]$ or $\propto \Pi^{-1}$ | **Precision and uncertainty**: precision is the inverse (co)variance or the confidence attached to a prediction or error; uncertainty is expressed either as the entropy of a belief or as inverse precision. |
| $F = E_{q(s)}[\ \ln q(s) - \ln P(o, s)\ ]$<br>$= D_{KL}[q(s) \parallel P(s)] - E_{q(s)}[\ln P(o \mid s)]$ | **Variational free energy (VFE)**: the free energy of current beliefs and a tractable upper bound on surprise (first row). It is assembled from surprise, the entropy of $q$, and the accuracy and complexity terms shown. Minimizing $F$ approximates Bayesian inference and governs inference about the present state of the system. |

| Expression | Description |
| --- | --- |
| $G(\pi) = \sum_{\tau} E_{q(o_\tau, s_\tau \mid \pi)}[\ \ln q(s_\tau \mid \pi) - \ln P(o_\tau, s_\tau)]$ <br> $\approx -E_q[\ D_{KL}[q(s_\tau \mid o_\tau, \pi) \parallel q(s_\tau \mid \pi)]\ ] - E_q[\ln P(o_\tau)]$ <br> $= D_{KL}[q(o_\tau \mid \pi) \parallel P(o_\tau)] \quad + \quad E_q[\ H[P(o_\tau \mid s_\tau)]\ ]$ | **Expected free energy (EFE)**: the free energy expected under a policy $\pi$. It splits into epistemic value (expected reduction in entropy / uncertainty) and pragmatic value (expected satisfaction of preferences); equivalently, risk + ambiguity. Surprise and entropy reappear here, in expectation over anticipated outcomes. $G$ governs the selection of action. |

## References

1. Clark, A.; Chalmers, D. The extended mind. *Analysis.* **1998**, *58*, 7–19.
2. Kirchhoff, M.; Kiverstein, J. *Extended Consciousness and Predictive Processing: A Third Wave View*; Routledge: London, UK; New York, NY, USA, 2019.
3. Menary, R. Cognitive integration and the extended mind. In *The Extended Mind*; Menary, R., Ed.; The MIT Press: Cambridge, MA, USA, 2010; pp. 227–243.
4. Sutton, J. Exograms and interdisciplinarity: History, the extended mind, and the civilizing process. In *The Extended Mind*; Menary, R., Ed.; The MIT Press: Cambridge, MA, USA, 2010; pp. 189–225.
5. Clark, A. *Surfing Uncertainty: Prediction, Action, and the Embodied Mind*; Oxford University Press: Oxford, UK, 2016.
6. Clark, A. *The Experience Machine: How Our Minds Predict and Shape Reality*; Penguin Books: London, UK, 2023.
7. Carl, M. Temporal dynamics of emotion and cognition in human translation: Integrating the Task Segment Framework and the HOF taxonomy. *Digital Studies in Language and Literature.* **2025**. https://doi.org/10.1515/dsll-2025-0002.
8. Kahneman, D. *Thinking, Fast and Slow*; Allen Lane: London, UK, 2011.
9. Evans, J.St.B.T.; Stanovich, K.E. Dual process theories of cognition: Advancing the debate. *Perspect. Psychol. Sci.* **2013**, *8*, 223–241.
10. Ajzen, I. The theory of planned behavior. *Organizational Behavior and Human Decision Processes.* **1991**, *50(2)*, 179–211.
11. Risku, H.; Rogl, R. Praxis and process meet halfway: The convergence of sociological and cognitive approaches in translation studies. *Transl. Interpreting.* **2022**, *14*, 32–49.
12. Muñoz Martín, R. Reembedding translation process research: An introduction. In *Reembedding Translation Process Research*; Muñoz Martín, R., Ed.; John Benjamins: Amsterdam, The Netherlands, 2016; pp. 1–20. https://doi.org/10.1075/btl.128.01mun.
13. Sannholm, R.; Risku, H. Situated minds and distributed systems in translation: Exploring the conceptual and empirical implications. *Target.* **2024**, *36*, 159–183.
14. Muñoz, Ricardo and Ana María Rojo López. (2018). Meaning. The Routledge Handbook of Translation and Cognition (edited by John W. Schwieter and Aline Ferreira).
15. Barsalou, L. W. (2010). Grounded cognition: Past, present, and future. Topics in Cognitive Science, 2(4), 716–724. https://doi.org/10.1111/j.1756-8765. 2010.01115.x
16. Banaruee, Hassan and Omid Khatin-Zadeh (2026) Embodied Language across Cultures. Palgrave. https://doi.org/10.1007/978-3-032-13848-4
17. Rietveld, E., & Kiverstein, J. (2014). A rich landscape of affordances. Ecological Psychology, 26, 325–352. doi: 10.1080/10407413.2014.958035
18. Kiverstein, J.D. and Rietveld, E. (2020). Scaling-up skilled intentionality to linguistic thought. Synthese. doi.org/10.1007/s11229-020-02540-3
19. Damasio, A.R. *Descartes' Error: Emotion, Reason, and the Human Brain*; Grosset/Putnam: New York, NY, USA, 1994.
20. Littau, K. Translation and the materialities of communication. *Transl. Stud.* **2016**, *9*, 82–96. https://doi.org/10.1080/14781700.2015.1063449.
21. O'Regan, J.K.; Noë, A. What it is like to see: A sensorimotor theory of perceptual experience. *Synthese.* **2001**, *129*, 79–103. https://doi.org/10.1023/A:1012699224677.

22. Seth, A.K. A predictive processing theory of sensorimotor contingencies: Explaining the puzzle of perceptual presence and its absence in synesthesia. *Cogn. Neurosci.* **2014**, *5*, 97–118. doi: 10.1080/17588928.2013.877880.
23. Ramstead, M.J.; Kirchhoff, M.D.; Friston, K.J. A tale of two densities: Active inference is enactive inference. *Adapt. Behav.* **2020**, *28*, 225–239.
24. Hurley, S.L. *Consciousness in Action*; Harvard University Press: Cambridge, MA, USA, 1998.
25. Hurley, S.L. The varieties of externalism. In *The Extended Mind*; Menary, R., Ed.; The MIT Press: Cambridge, MA, USA, 2010; pp. 101–154.
26. Ryle, G. *The Concept of Mind*; Hutchinson: London, UK, 1949.
27. Vervaeke, John ; Timothy P. Lillicrap andBlake A. Richards. (2012) Relevance Realization and the Emerging Framework in Cognitive Science, Journal of Logic and Computation, Volume 22, Issue 1 Pages 79–99, https://doi.org/10.1093/logcom/exp067
28. Bruineberg J. and Rietveld E. (2014) Self-organization, free energy minimization, and optimal grip on a field of affordances. Front Hum Neurosci. 2014 Aug 12;8:599. doi: 10.3389/fnhum.2014.00599. PMID: 25161615; PMCID: PMC4130179.
29. McCarthy, J., & Hayes, P. J. (1969). "Some philosophical problems from the standpoint of artificial intelligence." Machine Intelligence, 4, 463-502.
30. Harnad, Stevan (1990) The symbol grounding problem, Physica D: Nonlinear Phenomena, Volume 42, Issues 1–3, Pages 335-346, https://doi.org/10.1016/0167-2789(90)90087-6.
31. Friston, K. The free-energy principle: A unified brain theory? *Nat. Rev. Neurosci.* **2010**, *11*, 127–138.
32. Parr, T.; Benrimoh, D.A.; Vincent, P.; Friston, K.J. Precision and false perceptual inference. *Front. Integr. Neurosci.* **2018**, *12*, 39. doi: 10.3389/fnint.2018.00039.
33. Davies, S. The mathematics of mind-time. *Aeon.* **2025**. Available online: https://aeon.co/essays/consciousness-is-not-a-thing-but-a-process-of-inference (accessed on 21 July 2026).
34. Friston, K.J. Am I self-conscious? (Or does self-organization entail self-consciousness?). *Front. Psychol.* **2018**, *9*, 579. doi: 10.3389/fpsyg.2018.00579.
35. Kirchhoff, M.; Parr, T.; Palacios, E.; Friston, K.J.; Kiverstein, J. The Markov blankets of life: Autonomy, active inference and the free energy principle. *J. R. Soc. Interface.* **2018**, *15*, 20170792. doi: 10.1098/rsif.2017.0792.
36. Carl, M.; Mizowaki, T.; Raj, A.; Yamada, M.; Bandaru, D.S.; Ren, X. The behavioural translation style space: Towards simulating the temporal dynamics of affect, behaviour, and cognition in human translation production. *SKASE J. Transl. Interpreting.* **2025**, *18(2)*, 212–239. https://doi.org/10.33542/JTI2025-S-11.
37. Just, M. A., & Carpenter, P. A. (1980). A theory of reading: From eye fixations to comprehension. Psychological Review, 87(4), 329–354. doi.org
38. Rupert, Robert D. (2004). Challenges to the hypothesis of extended cognition. Journal of Philosophy 101 (8):389-428.
39. Parr, T.; Pezzulo, G.; Friston, K.J. *Active Inference: The Free Energy Principle in Mind, Brain, and Behavior*; The MIT Press: Cambridge, MA, USA, 2022. https://doi.org/10.7551/mitpress/12441.001.0001.
40. Garcia da Silva Oliveira, Deivide (2025). Rethinking Human-AI interactions through the maintainability: a new criterion for Trust and Glue in the Extended Mind Thesis. Trans/Form/Ação:Unesp Journal of Philosophy 48 (6):1-19.
41. Ramstead, M.J.D.; Wiese, W.; Miller, M.; Friston, K.J. Deep neurophenomenology: An active inference account of some features of conscious experience and of their disturbance in major depressive disorder. In *Expected Experiences*; Routledge: London, UK; New York, NY, USA, 2023; pp. 9–46.
42. Dragsted, B.; Carl, M. Towards a classification of translation styles based on eye-tracking and keylogging data. *J. Writ. Res.* **2013**, *5*, 133–158.
43. Mizowaki, T. Active inference in translation process: Validating cognitive model of translators' action selection. *Intercult. Commun. Rev.* Forthcoming.
44. Heins, C.; Millidge, B.; Demekas, D.; Klein, B.; Friston, K.J.; Couzin, I.D.; Tschantz, A. pymdp: A Python library for active inference in discrete state spaces. *J. Open Source Softw.* **2022**, *7*, 4098. https://doi.org/10.21105/joss.04098.
45. Chater, N.; Pickering, M.; Milward, D. What is incremental interpretation? *Edinb. Work. Pap. Cogn. Sci.* **1995**, *11*. Available online: https://courses.cit.cornell.edu/ling7710/readings/ChaterEtAlIncremental.pdf (accessed on 21 July 2026).
46. Hutchins, E. *Cognition in the Wild*; The MIT Press: Cambridge, MA, USA, 1995.
47. Rojo López, A.M.; Caldwell-Harris, C.L. Emotions in cognitive translation and interpreting studies. In *The Routledge Handbook of Translation, Interpreting and Bilingualism*; Ferreira, A., Schwieter, J.W., Eds.; Routledge: London, UK; New York, NY, USA, 2023; pp. 206–221.

48. Lodge, M.; Taber, C.S. The automaticity of affect for political leaders, groups, and issues: An experimental test of the hot cognition hypothesis. *Polit. Psychol.* **2005**, *26*, 455–482.
49. Rojo López, A.M.; Ramos Caro, M. The impact of translators' ideology on the translation process: A reaction time experiment. *MonTI: Monografías de traducción e interpretación*. **2014**, 247–271.
50. Naranjo, B.; Rojo López, A.M. The effects of musical (in)congruence on translation. *Target.* **2021**, *33*, 132–156.
51. Miljanović, Z.; Alves, F.; Brost, C.; Neumann, S. Directionality in translation: Throwing new light on an old question. *SKASE J. Transl. Interpreting* **2025**, *18*, 4–37. https://doi.org/10.33542/JTI2025-S-2.